\pdfoutput=1

\documentclass[11pt]{article}

\usepackage[final]{emnlp2021}

\usepackage{times}
\usepackage{latexsym}

\usepackage[T1]{fontenc}

\usepackage[utf8]{inputenc}

\usepackage{microtype}
\usepackage{amsmath,amssymb}
\usepackage{booktabs}
\usepackage{multirow}
\usepackage{graphicx}
\usepackage[table]{xcolor}

\usepackage[most]{tcolorbox}

\newtcolorbox{promptbox}[1][]{
    colback=gray!10,
    colframe=gray!30,
    coltitle=black,
    fonttitle=\bfseries,
    title={#1},
    boxrule=0.5pt,
    arc=4pt,
    left=6pt,
    right=6pt,
    top=6pt,
    bottom=6pt
}
\title{GUI-C$^2$: Coarse-to-Fine GUI Grounding via Difficulty-Aware Reinforcement Learning}

\author{Junlong Li\textsuperscript{†,\rm 1}, Chao Hao\textsuperscript{†,\rm 1}, Lap-Pui Chau\textsuperscript{\rm 1}, Yi Wang\textsuperscript{*,\rm 1}\\
\\
\textsuperscript{1}The Hong Kong Polytechnic University\\
\textsuperscript{†}\small Equal contribution,\textsuperscript{*}\small Corresponding author\\
\small \texttt{junlong.li@connect.polyu.hk}\\
\small Project page: \url{https://z1oong.github.io/GUI-C2/}
}

\begin{document}
\maketitle
\begin{abstract}
Existing agentic reinforcement learning methods for GUI grounding have limitations at two levels. At the data level, current approaches typically treat all training samples equally, although their training value to the baseline model varies with difficulty. Overlooking this can greatly reduce training efficiency or even cause collapse. At the strategy level, existing frameworks struggle to balance the trade-off between cropping larger regions for sufficient context and smaller ones for reduced redundancy, a tension inherent to tool-augmented grounding agents. In addition, overly complex decision-making is difficult for small-parameter models and significantly increases inference time. To address these issues, at the data level, we propose \textbf{GUI-D}, a data mining and difficulty scoring pipeline that identifies the training-worthy samples by proper testing and assigns difficulty scores to guide subsequent training weights. At the strategy level, we propose \textbf{GUI-C$^2$}, which employs an area-gated coarse-to-fine refinement mechanism that progressively narrows the visual field via model-internal uncertainty signals, adaptively reserving context for large targets while amplifying precision for small ones, reinforced by improvement-aware stage rewards that ensure each refinement genuinely advances grounding. Meanwhile, we simplify the decision-making process to greatly reduce additional inference time. Finally, extensive experiments show that our method achieves state-of-the-art performance. The code and data will be publicly available.
\end{abstract}

\section{Introduction}

\begin{figure}
    \centering
    \vspace{-5pt}
    \includegraphics[width=1\linewidth]{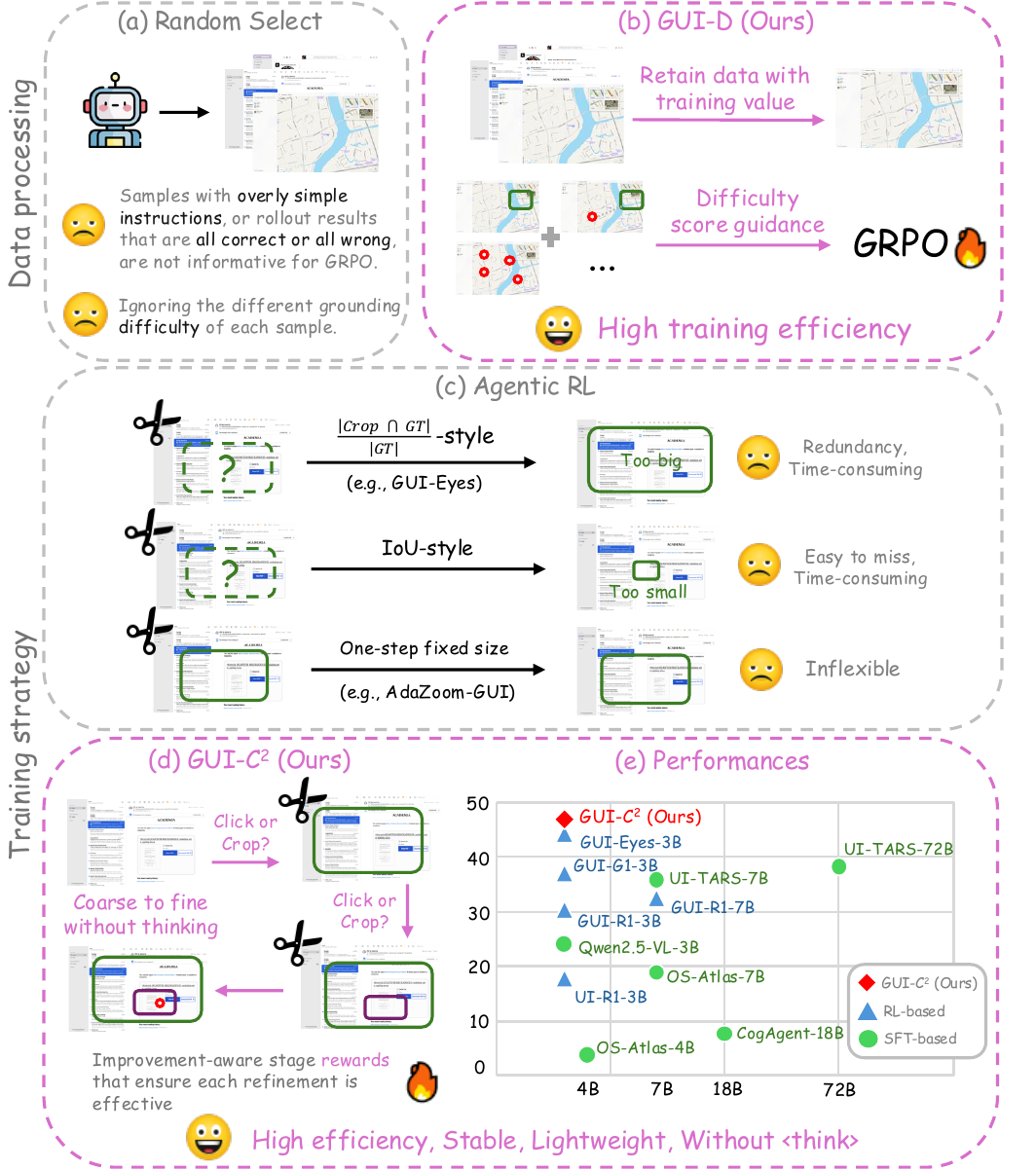}
    \caption{(a), (b), (c) and (d) highlight the limitations we target. (e) shows that GUI-C$^2$ achieves outstanding performance even with the 3B parameter setting.}
    \label{figtea}
    \vspace{-15pt}
\end{figure}

Graphical user interfaces (GUIs) \cite{cheng2024seeclickharnessingguigrounding} serve as the primary medium for human-computer interaction, making GUI grounding an indispensable foundational capability for autonomous language agents. Despite rapid progress in multimodal large language models (MLLMs) \cite{llava, bai2025qwen25vltechnicalreport, niu2025langtime, hao2025dynamic,li2026buildingegocentricproceduralai} and GUI agents \cite{androidworld}, accurate grounding remains challenging, especially on dense desktop interfaces, small icons, and visually similar widgets.

To address these visual challenges, equipping agents with a visual cropping tool has emerged as a promising strategy \cite{chen2026guieyestoolaugmentedperceptionvisual,tang2026uizoomeruncertaintydrivenadaptivezoomin,pei2026adazoomguiadaptivezoombasedgui}, enabling them to dynamically zoom in on target elements and mitigate visual redundancy. This naturally aligns with the capabilities of Reinforcement Learning (RL), particularly Group Relative Policy Optimization (GRPO) \cite{shao2024deepseekmathpushinglimitsmathematical}, which has demonstrated strong potential in training agentic tool-use behaviors. By formulating cropping as an autonomous action within an RL framework, agents can learn dynamic perception strategies tailored to varying interface complexities. However, as illustrated in Fig.~\ref{figtea}, existing agentic RL frameworks are fundamentally hindered by two critical bottlenecks \cite{zhao2026pointsguigguigroundingjourney} at the \textbf{data level} and the \textbf{strategy level}, respectively.

At the \textbf{data level}, current approaches \cite{pei2026adazoomguiadaptivezoombasedgui,chen2025guicoursegeneralvisionlanguage,gu2025uivenustechnicalreportbuilding,ye2025mobileagentv3fundamentalagentsgui} suffer from a "uniform treatment" bottleneck: they treat all training samples indiscriminately (Fig.~\ref{figtea}a). In reality, GUI grounding exhibits immense intrinsic variance in difficulty. Locating a prominent search bar requires minimal effort, whereas pinpointing a tiny, ambiguous icon within a dense cluster is highly challenging. Feeding all samples into the model with uniform training objectives leads to suboptimal policy learning. Easy samples tend to dominate the gradients while offering limited learning value; conversely, inappropriately handled hard samples fail to create meaningful intra-group differences during rollouts, which may even lead to training collapse. Mirroring human cognitive behavior, where individuals implicitly assess task difficulty and allocate more effort to harder problems, we propose \textbf{GUI-D}, a difficulty-aware data curation pipeline (Fig.~\ref{figtea}b). It filters out samples with low training value and leverages a comprehensive difficulty metric to dynamically assign weights during training, ensuring the model focuses its capacity on the most valuable data.

At the \textbf{strategy level}, existing frameworks struggle to balance context preservation and redundancy reduction when invoking cropping tools (Fig.~\ref{figtea}c). Some works \cite{pei2026adazoomguiadaptivezoombasedgui} adopt one-step fixed crop size, which is inflexible and fails to balance the required context with the element proportion. Applying the same crop ratio regardless of target scale leaves large targets over-cropped and small targets under-cropped (insufficient crop for precise localization), while also making precise grounding of tiny targets difficult. Other works \cite{chen2026guieyestoolaugmentedperceptionvisual} train models to autonomously select the crop area, but their reward designs tend to fall into two extremes: coverage-oriented rewards encourage overly conservative, large crops that preserve too much irrelevant context, while strict IoU-style rewards drive the predicted crop to become overly tight, where a small localization error also causes the crop to miss the target entirely. Furthermore, relying on complex, sequential decision-making (e.g., explicit thinking traces) significantly increases inference time and is often too difficult for small-parameter models. 

To overcome these strategy-level limitations, we propose \textbf{GUI-C$^2$}, an efficient agentic RL framework for coarse-to-fine GUI grounding (Fig.~\ref{figtea}d). Instead of a rigid, one-size-fits-all objective or time-consuming reasoning, GUI-C$^2$ elegantly simplifies the decision-making process into predicting the size of the target element (bounding box), which serves as both a refinement trigger and a dense supervision signal. This area-gated mechanism enables the model to autonomously initiate progressive visual refinement without explicit reasoning: an initial coarse crop preserves surrounding context for reliable target acquisition, while a subsequent fine crop, conditioned on the model's own uncertainty estimate, zooms into a tighter region for precise localization. Crucially, the multi-stage policy is stabilized by an improvement-aware refinement reward that explicitly incentivizes each crop to yield a more accurate grounding than the previous stage, preventing degenerative refinements and ensuring that additional computation translates into genuine gains. By coupling difficulty-aware data weighting with coarse-to-fine perception, our approach efficiently addresses the aforementioned grounding bottlenecks, achieving outstanding performance even with a 3B parameter setting (Fig.~\ref{figtea}e). Our contributions can be summarized as follows:

\begin{itemize}
    \item We propose GUI-D, a data filtering and dynamic weighting pipeline for GRPO that leverages the difficulty metric to optimize sample utilization, improving training efficiency.
    \item We introduce GUI-C$^2$, an efficient agentic RL framework for coarse-to-fine GUI grounding. By formulating multi-stage cropping as bounding box prediction, it applies distinct learning behaviors to simple and hard samples, reducing decision complexity while improving inference efficiency.
    \item By combining dynamic difficulty weighting and coarse-to-fine cropping, our method achieves state-of-the-art performance under comparable conditions (3B) on three commonly used benchmarks, remarkably using only 4,624 training samples.
\end{itemize}

\begin{figure*}
    \centering
        \vspace{-10pt}
    \includegraphics[width=1\linewidth]{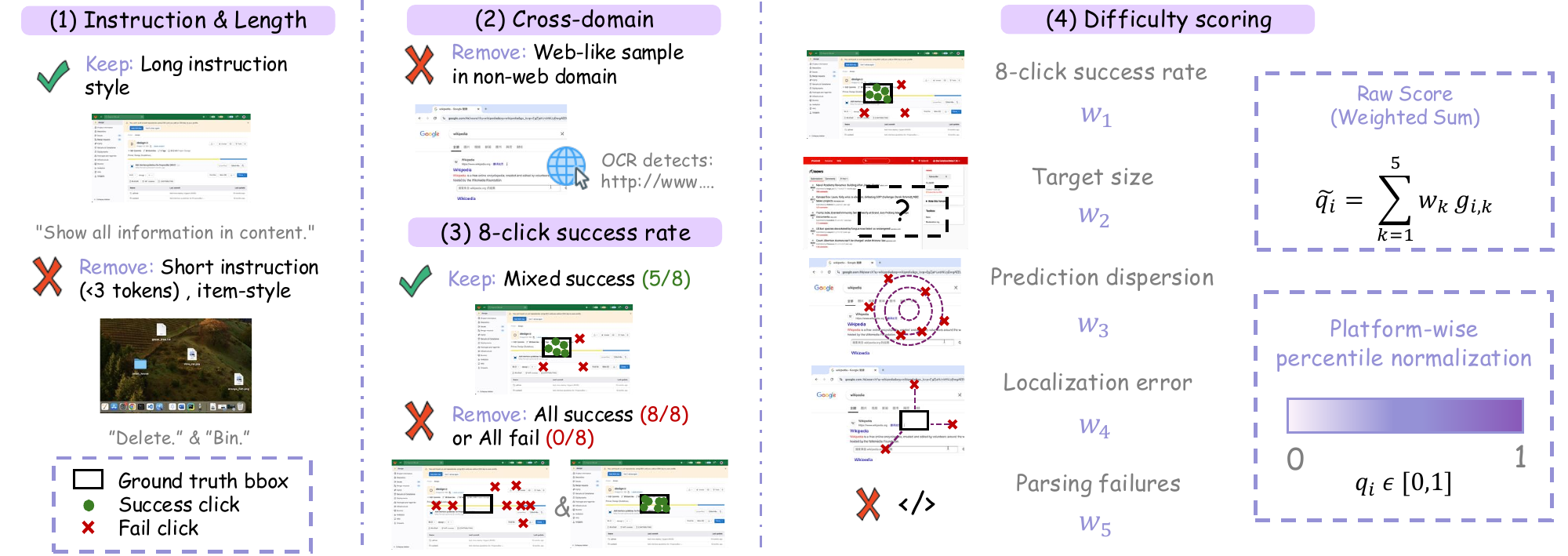}
    \caption{Overview of GUI-D. (1), (2), and (3) illustrate the data filtering pipeline, while (4) shows the main factors considered in our difficulty scoring.}
    \label{figguid}
    \vspace{-16pt}
\end{figure*}

\section{Related Work}


\textbf{GUI Grounding.}
Existing grounding methods can be broadly divided into three paradigms. The first line relies on test-time refinement \cite{zhang2026bamitrainingfreebiasmitigation,tang2026uizoomeruncertaintydrivenadaptivezoomin,wu2025dimoguiadvancingtesttimescaling,hao2025uncertaintyawareguiagentadaptive, hao2026seg}, which uses complex test-time rules to achieve higher click accuracy without training. The second line employs supervised fine-tuning (SFT) \cite{wu2024osatlasfoundationactionmodel,hong2024cogagentvisuallanguagemodel,qin2025uitarspioneeringautomatedgui} to enable models to substantially improve performance through imitation. However, both methods have notable limitations. Test-time methods sacrifice inference time, yet GUI grounding is a fundamental action for GUI agents, and long-horizon tasks involve multiple clicks. Applying expensive, time-consuming test-time rules to a simple click action is highly inefficient and unscalable, while training-free methods often yield only marginal performance gains. SFT methods, on the other hand, require large amounts of annotated data and tend to lack strong generalization. The third line, reinforcement learning \cite{chen2026guieyestoolaugmentedperceptionvisual,pei2026adazoomguiadaptivezoombasedgui,zhou2025guig1understandingr1zeroliketraining}, endows models with stronger generalization capabilities and has become the current mainstream trend. 


\textbf{GRPO and Data Setting}.
With the introduction of the GRPO-based framework by DeepSeek-R1 \cite{guo2025}, this efficient and more flexible strategy was soon extended to the GUI agent domain. InfiGUI-R1 \cite{liu2025infiguir1advancingmultimodalgui} began encouraging models to think before acting, whereas GUI-G1 \cite{zhou2025guig1understandingr1zeroliketraining} demonstrated through experiments that the thinking process is not essential for the fundamental action like grounding. Although these methods effectively utilize reinforcement learning, they still rely on pure text-based reasoning, overlooking visual information. Subsequently, some works \cite{chen2026guieyestoolaugmentedperceptionvisual} began exploring agentic RL solutions, training models to proactively invoke tools to improve grounding accuracy for fine-grained elements. However, on the data side, current works largely ignore the guiding value of data difficulty for the baseline model. On the strategy side, existing methods fail to strike a balance among policy complexity, tool invocation configuration, and inference time.

\begin{figure*}
    \centering
  \vspace{-4pt}
    \includegraphics[width=1\linewidth]{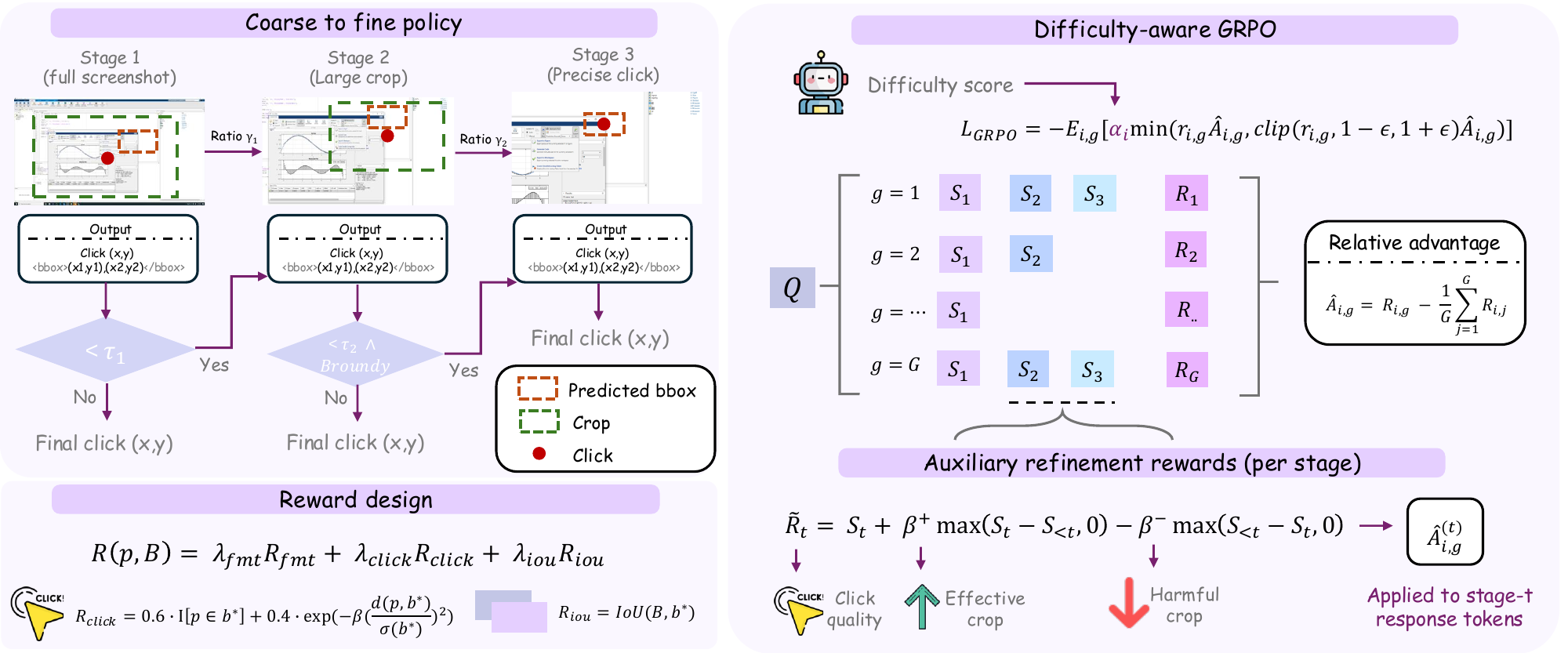}
    \caption{Overall framework of GUI-C². The left panel illustrates the coarse-to-fine policy and our overall reward function design. The right panel shows how the difficulty score is introduced for adjustment during training, and for each tool invocation stage, we introduce additional rewards to encourage effective refinement.}
    \label{figguic}
    \vspace{-10pt}
\end{figure*}

\section{GUI-D: Difficulty-Aware Data Curation}

In this section, we introduce \textbf{GUI-D}, a lightweight data curation pipeline that removes low-value GUI grounding samples and assigns each remaining sample a difficulty score for later GRPO training.

\subsection{Data Filtering Pipeline}

As shown in Fig.~\ref{figguid}, following the data processing workflow of previous work \cite{zhou2025guig1understandingr1zeroliketraining}, we first remove simple item-name prompts (e.g., UI element identifiers) and retain only action-oriented instructions, specifically, samples whose first token is a verb and contain at least three tokens. This filters out superficial lexical-matching queries and favors samples that require genuine semantic understanding. Then we address cross-domain contamination: we apply OCR to detect web-specific patterns (URLs, "www") in samples outside the web domain and remove them, thereby enforcing cleaner domain separation between desktop, mobile, and web subsets.

After that, we filter samples by rollout informativeness. GRPO improves a policy by increasing the probability of responses with higher rewards among multiple rollouts, samples that are always solved or always failed provide limited training signal. Thus, for each remaining sample, we simulate $K=8$ click rollouts and we keep only samples with mixed outcomes.

\subsection{Difficulty Definition}

To more reasonably reflect the true clicking difficulty of a sample for the baseline model, and to ensure that this reflection can indeed guide and optimize training, GUI-D assigns each sample a scalar difficulty score, providing an interpretable measure of grounding hardness based on observable rollout behavior and visual target properties. Existing work \cite{wang2026guiperturbeddomainrandomizationreveals,liu2026zoomessencetrainlessgui} shows that the grounding difficulty of a sample for a model is not solely reflected in the intuitively observable ground-truth area. The 8-click success rate most directly indicates the difficulty of a sample and accordingly receives the largest weight in the final score. However, relying solely on this factor fails to capture fine-grained difficulty differences between samples. Therefore, in addition to success rate and target size, we further incorporate localization error, prediction dispersion, and parsing failures. Even if none of the clicks hit the target, a sample with clicks that are close to the ground truth is easier. Thus, localization error measures the average distance of the 8 clicks to ground truth. Moreover, when the 8 clicks are highly dispersed, it indicates excessive distractors in the sample; hence, we introduce prediction dispersion, which represents 8-click variance. Finally, parsing failure reflects whether the model outputs a legally parsable format and accordingly receives the smallest weight in the final difficulty measure.

We summarize these factors with a weighted raw difficulty score. Because GUI platforms have different visual layouts and element scales, we normalize difficulty scores within each platform by percentile ranking $q_i\in[0,1]$. More details can be found in Appendix~\ref{apa}.

\section{GUI-C$^2$: Coarse-to-Fine Strategy}

Given a screenshot $I$ and an instruction $q$, the goal is to predict a click point $p=(x,y)$ that falls inside the target bounding box $b^{\star}$. Unlike reasoning-heavy GUI agents, GUI-C$^2$ directly outputs an executable action line without an explicit thinking trace, and simplifies the decision of whether to perform a crop action into a bbox area prediction. This design keeps the action space compact, and improves inference efficiency and stability.

\subsection{Overall Architecture}

As shown in Fig.~\ref{figguic}, GUI-C$^2$ formulates GUI grounding as a staged visual refinement process. At each stage $t$, the policy $\pi_{\theta}$ observes a view $I_t$ and outputs both a click point and a bounding box:
\begin{equation}
\small
(p_t, B_t) \sim \pi_{\theta}(I_t, q),
\qquad
B_t=(x_1^t,y_1^t,x_2^t,y_2^t).
\end{equation}
The click point is used as the executable action, while the predicted box is used as a lightweight visual uncertainty signal to decide whether further refinement is needed.

The first stage operates on the full screenshot: $I_1 = I$. If the predicted target is visually small, GUI-C$^2$ crops a local view around the predicted click and performs a second grounding stage. A further crop can be triggered in the same manner. Let $W_t,H_t$ be the width and height of the current view. We define the normalized predicted area as
\begin{equation}
\small
A(B_t; I_t)
=
\frac{(x_2^t-x_1^t)(y_2^t-y_1^t)}{W_tH_t}.
\end{equation}
The first refinement is triggered by
\begin{equation}
\small
z_1 = \mathbb{I}\left[A(B_1; I_1)<\tau_1\right],
\end{equation}
and the second refinement is triggered by
\begin{equation}
\small
z_2 = \mathbb{I}\left[A(B_2; I_2)<\tau_2 \;\lor\; \mathrm{Boundary}(B_2,I_2)\right].
\end{equation}
Here $\tau_1$ and $\tau_2$ are area thresholds, and $\mathrm{Boundary}(B_2,I_2)$ indicates that the predicted box touches the boundary of the current crop. The thresholds control when to crop, while the crop ratios control how much visual context to keep.

Given a crop ratio $\gamma_t\in(0,1]$, the next view is centered at the predicted click point $p_t$:
\begin{equation}
\small
I_{t+1}
=
\mathrm{Crop}\left(I_t, p_t, \gamma_t W_t, \gamma_t H_t\right).
\end{equation}
Thus, GUI-C$^2$ uses two independent sets of parameters: \{$\tau_1$,$\tau_2$\} for refinement triggering, \{$\gamma_1$,$\gamma_2$\} for crop scale control. At inference time, the model follows this deterministic gating rule and executes the deepest valid prediction. No candidate reranking or chain-of-thought decoding is used.

\subsection{Area-Guided Coarse-to-Fine Refinement}

The area gate is motivated by the observation that small GUI targets are harder to localize from a full screenshot. Directly selecting a point from the full screenshot forces the model to solve both global semantic search and fine-grained localization simultaneously. GUI-C$^2$ decomposes this problem into a coarse-to-fine process: the first stage finds a rough target region, and subsequent stages localize the target within a simplified local context.

Using the predicted area as the trigger has two advantages. First, it provides a model-internal uncertainty proxy without requiring an extra confidence head or further reasoning, the model always learns the same primitive action: predict a click and a box. Second, it avoids unnecessary crops for already visible targets, preserving global context when it is useful. The boundary condition in the second stage further handles cases where the target may be truncated by the previous crop.

The asymmetric thresholds $\tau_1$ and $\tau_2$ reflect the different roles of the two stages. The first prediction is made on the full screenshot, where clutter is strongest; therefore, the first gate should be relatively permissive. After one crop, the target is already magnified, so the second gate can be more conservative to avoid over-cropping and losing context. In the experiments, these parameters are instantiated in the hyperparameter study rather than treated as fixed constants in the method definition.

\subsection{Difficulty-Modulated GRPO Training}

GUI-C$^2$ is trained with GRPO on the curated GUI-D data. The previous section defines a difficulty score $q_i\in[0,1]$ for each training sample. We use this score to softly modulate the contribution of each example instead of changing the model architecture. The difficulty multiplier is
\begin{equation}
\small
\alpha_i
=
\mathrm{clip}_{[\alpha_{\min},\alpha_{\max}]}
\left(1+\lambda_d(q_i-0.5)\right),
\end{equation}
where $\lambda_d$ controls the strength of difficulty weighting. This design mildly up-weights informative hard samples while preventing noisy examples from dominating training.

For each instruction, the model samples a group of $G$ completions. Let $R_{i,g}$ be the reward of the $g$-th completion for sample $i$. GRPO estimates the relative advantage within the sampled group:
\begin{equation}
\small
\hat{A}_{i,g}
=
R_{i,g}
-
\frac{1}{G}\sum_{j=1}^{G}R_{i,j}.
\end{equation}
The policy objective is then weighted by the GUI-D difficulty multiplier:
\begin{equation}
\small
\begin{split}
\mathcal{L}_{\mathrm{GRPO}}
=&
-
\mathbb{E}_{i,g}
\Bigg[
\alpha_i
\min\Bigg(
r_{i,g}\hat{A}_{i,g},
\Bigg.
\\
&\Bigg.
\mathrm{clip}(r_{i,g},1-\epsilon,1+\epsilon)\hat{A}_{i,g}
\Bigg)
\Bigg],
\end{split}
\end{equation}
where
\begin{equation}
\small
r_{i,g}
=
\frac{\pi_{\theta}(a_{i,g}\mid I_i,q_i)}
{\pi_{\theta_{\mathrm{old}}}(a_{i,g}\mid I_i,q_i)}.
\end{equation}
This weighting connects data curation and policy optimization: GUI-D identifies samples with useful training signal, and GUI-C$^2$ converts that signal into a soft optimization bias.

\begin{table*}[t]
\centering
\vspace{-11pt}
\scalebox{0.79}{
\small
\setlength{\tabcolsep}{5.5pt}
\begin{tabular}{lccccccccccccccc}
\toprule
\multirow{2}{*}{\textbf{Model}} & 
\multicolumn{2}{c}{\textbf{CAD}} & 
\multicolumn{2}{c}{\textbf{Development}} & 
\multicolumn{2}{c}{\textbf{Creative}} & 
\multicolumn{2}{c}{\textbf{Scientific}} & 
\multicolumn{2}{c}{\textbf{Office}} & 
\multicolumn{2}{c}{\textbf{OS}} & 
\multicolumn{3}{c}{\textbf{Avg.}} \\
\cmidrule(lr){2-3}   
\cmidrule(lr){4-5} 
\cmidrule(lr){6-7}
\cmidrule(lr){8-9}
\cmidrule(lr){10-11}
\cmidrule(lr){12-13}
\cmidrule(lr){14-16}
 & Text & Icon & 
   Text & Icon & 
   Text & Icon & 
   Text & Icon & 
   Text & Icon & 
   Text & Icon & 
   Text & Icon & Avg \\
\midrule
\rowcolor{gray!20}\multicolumn{16}{l}{\textbf{Proprietary Models}} \\
GPT-4o \cite{openai2024gpt4ocard} & 2.0 & 0.0 & 1.3 & 0.0 & 1.0 & 0.0 & 2.1 & 0.0 & 1.1 & 0.0 & 0.0 & 0.0 & 1.3 & 0.0 & 0.8 \\
Claude Computer Use \cite{cluade} & 14.5 & 3.7 & 22.0 & 3.9 & 25.9 & 3.4 & 33.9 & 15.8 & 30.1 & 16.3 & 11.0 & 4.5 & 23.4 & 7.1 & 17.1 \\
\midrule
\rowcolor{gray!20}\multicolumn{16}{l}{\textbf{General Open-source Models}} \\
Qwen2.5-VL-3B \cite{bai2025qwen25vltechnicalreport} & 9.1 & 7.3 & 22.1 & 1.4 & 26.8 & 2.1 & 38.2 & 7.3 & 33.9 & 15.1 & 10.3 & 1.1 & 23.6 & 3.8 & 16.1 \\
Qwen2.5-VL-7B \cite{bai2025qwen25vltechnicalreport} & 16.8 & 1.6 & 46.8 & 4.1 & 35.9 & 7.7 & 49.3 & 7.3 & 52.5 & 20.8 & 37.4 & 6.7 & 38.9 & 7.1 & 26.8 \\
\midrule
\rowcolor{gray!20}\multicolumn{16}{l}{\textbf{GUI-Specific Models(SFT(+RL))}} \\
CogAgent-18B \cite{hong2024cogagentvisuallanguagemodel} & 7.1 & 3.1 & 14.9 & 0.7 & 9.6 & 0.0 & 22.2 & 1.8 & 13.0 & 0.0 & 5.6 & 0.0 & 12.0 & 0.8 & 7.7 \\
OS-Atlas-7B \cite{wu2024osatlasfoundationactionmodel} & 12.2 & 4.7 & 33.1 & 1.4 & 28.8 & 2.8 & 37.5 & 7.3 & 33.9 & 5.7 & 27.1 & 4.5 & 28.1 & 4.0 & 18.9 \\
ShowUI-2B \cite{lin2024showuivisionlanguageactionmodelgui} & 2.5 & 0.0 & 16.9 & 1.4 & 9.1 & 0.0 & 13.2 & 7.3 & 15.3 & 7.5 & 10.3 & 2.2 & 10.8 & 2.6 & 7.7 \\
UGround-7B \cite{gou2025navigatingdigitalworldhumans} & 14.2 & 1.6 & 26.6 & 2.1 & 27.3 & 2.8 & 31.9 & 2.7 & 31.6 & 11.3 & 17.8 & 0.0 & 25.0 & 2.8 & 16.5 \\
UGround-V1-7B \cite{gou2025navigatingdigitalworldhumans} & 15.8 & 1.2 & 51.9 & 2.8 & \underline{47.5} & 9.7 & 57.6 & 14.5 & 60.5 & 13.2 & 38.3 & 7.9 & 45.2 & 8.1 & 31.1 \\
UI-TARS-2B \cite{qin2025uitarspioneeringautomatedgui} & 17.8 & 4.7 & 47.4 & 4.1 & 42.9 & 6.3 & 56.9 & 17.3 & 50.3 & 17.0 & 21.5 & 5.6 & 39.6 & 8.4 & 27.7 \\
UI-TARS-7B \cite{qin2025uitarspioneeringautomatedgui} & 20.8 & 9.4 & 58.4 & \underline{12.4} & 50.0 & 9.1 & 63.9 & \textbf{31.8} & 63.3 & 20.8 & 30.8 & 16.9 & 47.8 & 16.2 & 35.7 \\
InfiGUI-R1-3B \cite{liu2025infiguir1advancingmultimodalgui} & 33.0 & \underline{14.1} & 51.3 & \underline{12.4} & 44.9 & 7.0 & 58.3 & 20.0 & 65.5 & 28.3 & 43.9 & 12.4 & 49.1 & 14.1 & 35.7 \\
\midrule
\rowcolor{gray!20}\multicolumn{16}{l}{\textbf{GUI-Specific Models(RL Only)}} \\
UI-R1-3B \cite{lu2025uir1enhancingefficientaction} & 11.2 & 6.3 & 22.7 & 4.1 & 27.3 & 3.5 & 42.4 & 11.8 & 32.2 & 11.3 & 13.1 & 4.5 & 24.9 & 6.4 & 17.8 \\
GUI-R1-3B \cite{luo2025guir1generalistr1style} & 26.4 & 7.8 & 33.8 & 4.8 & 40.9 & 5.6 & 61.8 & 17.3 & 53.6 & 17.0 & 28.1 & 5.6 & 45.1 & 8.1 & 30.2 \\
GUI-R1-7B \cite{luo2025guir1generalistr1style} & 23.9 & 6.3 & 49.4 & 4.8 & 38.9 & 8.4 & 55.6 & 11.8 & 58.7 & 26.4 & 42.1 & 16.9 & 45.9 & 11.2 & 32.4 \\
SE-GUI-3B \cite{yuan2025enhancingvisualgroundinggui} & 38.1 & 12.5 & 55.8 & 7.6 & 47.0 & 4.9 & 61.8 & 16.4 & 59.9 & 24.5 & 40.2 & 12.4 & 50.4 & 11.8 & 35.9 \\
GUI-G1-3B \cite{zhou2025guig1understandingr1zeroliketraining} & 39.6 & 9.4 & 50.7 & 10.3 & 36.6 & 11.9 & 61.8 & \underline{30.0} & 67.2 & \underline{32.1} & 23.5 & 10.6 & 49.5 & \underline{16.8}& 37.1 \\
GUI-Eyes-3B \cite{chen2026guieyestoolaugmentedperceptionvisual} & \textbf{48.2} & 9.4 & \underline{70.8} & \underline{12.4} & \textbf{56.6} & \underline{13.3} & \textbf{69.4} & 19.1 & \textbf{75.7} & 24.5 & \textbf{59.8} & \underline{20.2} & \textbf{62.8} & 15.7 & 44.8
\\
\midrule
\textbf{GUI-C$^2$-3B(ours)} & \underline{43.7} & \cellcolor{blue!15}\textbf{21.9} & \cellcolor{blue!15}\textbf{72.1} & \cellcolor{blue!15}\textbf{19.3} & \cellcolor{blue!15}\textbf{56.6} & \cellcolor{blue!15}\textbf{14.7} & \underline{68.8} & 21.8 & \underline{74.6} & \cellcolor{blue!15}\textbf{37.7} & \underline{57.9} & \cellcolor{blue!15}\textbf{27.0} & \underline{61.6} & \cellcolor{blue!15}\textbf{21.7} & \cellcolor{blue!15}\textbf{46.4} 
\\
\bottomrule
\end{tabular}
}
\vspace{-4pt}
\caption{Performance comparison of different agent models on ScreenSpot-Pro. Results marked in \textbf{bold} and \underline{underlined} represent the best and second-best performance. Purple highlights mark the categories in which our method achieves the best performance.}
\label{tab:screenspot-pro}
\vspace{-17pt}
\end{table*}

\subsection{Reward Design}

The reward function is designed to jointly encourage executable formatting, accurate clicking, and spatially meaningful bounding boxes. For a predicted action $(p,B)$ and ground truth box $b^{\star}$, we use
\begin{equation}
\small
R(p,B)
=
\lambda_{\mathrm{fmt}}R_{\mathrm{fmt}}
+
\lambda_{\mathrm{click}}R_{\mathrm{click}}
+
\lambda_{\mathrm{iou}}R_{\mathrm{iou}}.
\end{equation}
The format reward is
\begin{equation}
\small
R_{\mathrm{fmt}}
=
\mathbb{I}\left[p \neq \varnothing \;\land\; B\neq \varnothing\right],
\end{equation}
which stabilizes action parsing. The click reward measures whether the predicted click falls inside the target and how far it is:
\begin{equation}
\small
R_{\mathrm{click}} = 0.6 \cdot \mathbb{I}[p \in b^\star] + 0.4 \cdot \exp\left( -\beta \left( \frac{d(p, b^\star)}{\sigma(b^\star)} \right)^2 \right).
\end{equation}
The box reward provides dense spatial supervision through IoU:
\begin{equation}
\small
R_{\mathrm{iou}}
=
\mathrm{IoU}(B,b^{\star}).
\end{equation}
The click reward directly optimizes the evaluation target, while the IoU term shapes the auxiliary bounding-box prediction used by the refinement gate. The format term is kept lightweight because formatting is necessary for reliable execution but should not dominate grounding accuracy.

In addition to the base action reward, GUI-C$^2$ uses auxiliary stage rewards to train the refinement outputs. These rewards are not added to the final evaluation reward; instead, they define separate GRPO advantages for the tokens generated at each refinement stage. Let $S_t$ be the click quality score (click reward) at stage $t$, and let $S_{<t}=\max_{j<t}S_j$ be the best previous score. The auxiliary refinement reward is
\begin{equation}
\small
\begin{split}
\widetilde{R}_t
&=
S_t
+
\beta^{+}\max(S_t-S_{<t},0)
\\
&-
\beta^{-}\max(S_{<t}-S_t,0),
\end{split}
\end{equation}
where $\beta^{+}$ rewards improvement and $\beta^{-}$ penalizes harmful refinement. This encourages genuinely helpful refinement and aligns the multi-stage policy with the final click objective. The refinement advantage is computed within the same stage:
\begin{equation}
\small
\hat{A}^{(t)}_{i,g}
=
\widetilde{R}^{(t)}_{i,g}
-
\frac{1}{G}\sum_{j=1}^{G}\widetilde{R}^{(t)}_{i,j}.
\end{equation}
The policy loss is then applied to the corresponding stage-$t$ response tokens. Therefore, refinement rewards affect training through stage-specific advantages rather than by changing the final click reward directly.

\begin{table*}[t]
\centering
\vspace{-8pt}
\scalebox{0.91}{
\small
\begin{tabular}{l c | ccc c | ccc c}
\toprule
\multirow{2}{*}{\textbf{Model}} & \multirow{2}{*}{\textbf{Training Samples}} & \multicolumn{4}{c|}{\textbf{ScreenSpot Accuracy (\%)}} & \multicolumn{4}{c}{\textbf{ScreenSpot-v2 Accuracy (\%)}} \\
& & Mobile & Desktop & Web & Avg. & Mobile & Desktop & Web & Avg. \\
\midrule
\rowcolor{gray!20}\multicolumn{10}{l}{\textbf{Proprietary Models}} \\
GPT-4o \cite{openai2024gpt4ocard} & - & 21.9 & 17.8 & 9.4 & 18.8 & 22.5 & 22.2 & 12.4 & 20.1 \\
\midrule
\rowcolor{gray!20}\multicolumn{10}{l}{\textbf{General Open-source Models}} \\
Qwen2-VL-7B \cite{wang2024qwen2vlenhancingvisionlanguagemodels} & - & 50.3 & 40.4 & 27.4 & 42.9 & 39.4 & 50.1 & 27.7 & 39.8 \\
Qwen2.5-VL-3B \cite{bai2025qwen25vltechnicalreport} & - & - & - & - & 55.5 & 55.5 & 44.0 & 39.1 & 46.9 \\
Qwen2.5-VL-7B \cite{bai2025qwen25vltechnicalreport} & - & - & - & - & \underline{84.7} & \textbf{92.8} & 78.4 & \underline{85.4} & 86.5 \\
\midrule
\rowcolor{gray!20}\multicolumn{10}{l}{\textbf{GUI-Specific Models}} \\
CogAgent-18B \cite{hong2024cogagentvisuallanguagemodel} & 222M & 57.8 & 31.6 & 40.1 & 47.4 & 50.6 & 51.6 & 54.1 & 52.8 \\
SeeClick-7B \cite{cheng2024seeclickharnessingguigrounding} & 1M & 68.1 & 48.8 & 41.8 & 53.4 & 51.8 & 65.5 & 40.7 & 53.9 \\
UGround-7B \cite{gou2025navigatingdigitalworldhumans} & 10M & 75.9 & 75.8 & 78.3 & 73.3 & 74.3 & 74.9 & 78.6 & 76.3 \\
ShowUI-2B \cite{lin2024showuivisionlanguageactionmodelgui} & 256K & 84.8 & 70.8 & 76.2 & 75.1 & 70.0 & 85.1 & 73.3 & 77.3 \\
OSAtlas-4B \cite{wu2024osatlasfoundationactionmodel} & 13M & 56.2 & 74.9 & 69.9 & 68.5 & 74.9 & 56.9 & 70.7 & 68.5 \\
OSAtlas-7B \cite{wu2024osatlasfoundationactionmodel} & 13M & 85.0 & 78.8 & 84.5 & 82.5 & 78.3 & 85.5 & 83.8 & 83.3 \\
Aguvis-7B \cite{xu2025aguvisunifiedpurevision} & 1M & \textbf{86.9} & \underline{82.4} & \underline{84.7} & 84.4 & \underline{89.6} & \underline{86.8} & 84.9 & \underline{87.3} \\
UI-TARS-2B \cite{qin2025uitarspioneeringautomatedgui} & 2M & 85.0 & 81.4 & 79.8 & 82.3 & 87.9 & 81.4 & 82.9 & 84.7 \\
\midrule
\textbf{GUI-C$^2$-3B (Ours)} & 4.6K & \underline{85.5} & \cellcolor{blue!15}\textbf{87.1} & \cellcolor{blue!15}\textbf{85.1} & \cellcolor{blue!15}\textbf{85.8} & 88.2 & \cellcolor{blue!15}\textbf{88.6} & \cellcolor{blue!15}\textbf{86.7} & \cellcolor{blue!15}\textbf{87.8} \\
\bottomrule
\end{tabular}
}
\vspace{-6pt}
\caption{Comparison of model performance on ScreenSpot and ScreenSpot-v2. Results marked in \textbf{bold} and \underline{underlined} represent the best and second-best performance. Purple highlights mark the categories in which our method achieves the best performance.}
\label{tab:screenspot-full}
\vspace{-12pt}
\end{table*}
\begin{table}
    \centering
    \vspace{-5pt}
    \scalebox{0.70}{
    \begin{tabular}{lc}
    \hline  
         \textbf{Model}& \textbf{ScreenSpot-Pro Avg.}\\
         \midrule
            Qwen2.5-VL-7B \cite{bai2025qwen25vltechnicalreport} & 26.8\\
        GUI-R1-7B \cite{luo2025guir1generalistr1style} & 32.4\\
        JEDI-7B \cite{xie2025scalingcomputerusegroundinguser}& 39.5\\
        GUI-Actor-7B \cite{wu2025guiactorcoordinatefreevisualgrounding} & 44.6\\
        SE-GUI-7B \cite{yuan2025enhancingvisualgroundinggui} & 47.3\\
        GUI-G$^2$-7B \cite{tang2025guig2gaussianrewardmodeling} & 47.5\\
        OpenCUA-7B \cite{wang2025opencuaopenfoundationscomputeruse} & 50.0\\
        GTA1-7B \cite{yang2025gta1guitesttimescaling} & \underline{50.1}\\
        \rowcolor{gray!20} \textbf{GUI-C$^2$-7B (Ours)} & \textbf{50.8}\\
         \bottomrule
    \end{tabular}
    
}  
\vspace{-6pt}
\caption{Performance comparison of GUI-C$^2$-7B. All methods use Qwen2/2.5-VL-7B as the base model.}
    \label{tabpro7b}
    \vspace{-8pt}
\end{table}
\begin{table}
    \centering
    \resizebox{0.8\linewidth}{!}{
    \begin{tabular}{lcc}
    \hline
         &\textbf{ Inference time (s) }& \textbf{Avg.}\\
     \midrule
    GUI-C$^2$ & 3.05 & 46.4\\
       w/o tool use & 1.50 & 37.0\\
         w/o coarse-to-fine policy& 2.42 & 41.4\\
        w/o difficulty-aware & 3.05 & 43.7\\
        w/ self-action-decide  & 24.81 & 43.5\\
        adaptive crop ratio & 3.05 & 43.8\\
        w/ self-action-ratio-decide & 32.71 & 43.4\\
        \hline
    \end{tabular}}
\vspace{-4pt}
    \caption{Ablation experiments on ScreenSpot-Pro.}
    \label{tabab}
    \vspace{-19pt}
\end{table}

\section{Experiments}
\label{sec:experiments}

\subsection{Implementation Details}
\label{subsec:implementation_details}

\textbf{Training Details.} Following previous works \cite{chen2026guieyestoolaugmentedperceptionvisual,zhou2025guig1understandingr1zeroliketraining}, we choose Qwen2.5-VL-3B/7B-Instruct \cite{bai2025qwen25vltechnicalreport} as our base model, and trained using the VLM-R1 \cite{shen2025vlmr1stablegeneralizabler1style} framework, ensuring a clean comparison with other training strategies. The coefficients in the reward function are set to $\lambda_{\mathrm{fmt}}=0.1$, $\lambda_{\mathrm{click}}=0.6$ and $\lambda_{\mathrm{iou}}=0.3$. We conduct training for one epoch on 4 NVIDIA H20 GPUs about 17 hours for 3B and 26 hours for 7B, with a global batch size of 32 and a learning rate of $1\times10^{-6}$, bf16 precision. We set temperature to $1.0$ and KL coefficient to $0$. More details are in Appendix~\ref{apc}.

\textbf{Training Dataset and Evaluation.} We use GUI-D to construct a 4624 (3B)/4824 (7B)-sample dataset spanning three domains: \textbf{Mobile }(from UI-BERT \cite{bai2021uibertlearninggenericmultimodal}), \textbf{Web }(from OS-Atlas \cite{wu2024osatlasfoundationactionmodel}), and \textbf{Desktop}(from OS-Atlas, covering Windows, Linux, and MacOS). For comprehensive evaluation, we select three widely used benchmarks, including ScreenSpot-Pro \cite{li2025screenspotproguigroundingprofessional}, ScreenSpot \cite{cheng2024seeclickharnessingguigrounding}, ScreenSpot-V2. All baseline results are collected from publicly released resources. All experiments not otherwise specified are conducted using GUI-C$^2$-3B.

\subsection{Experimental Results}

As shown in Table~\ref{tab:screenspot-pro}, GUI-C$^2$-3B achieves state-of-the-art (SOTA) performance under comparable conditions on the challenging ScreenSpot-Pro benchmark. With only 4,624 training samples, it attains an overall accuracy of 46.4\%, significantly outperforming GUI-G1 which requires 17,000 samples. Meanwhile, it surpasses the agentic method GUI-Eyes that relies on a thinking process, achieving fast and accurate clicking without explicit reasoning. GUI-C$^2$-3B achieves the best performance on seven fine-grained categories like CAD icons and Development text/icon samples. Notably, GUI-C$^2$-3B demonstrates stronger grounding accuracy on icon-type targets than on text. Across the six major categories, it achieves the best icon-type performance on all except Scientific. Among all fine-grained categories, relative to the baseline model Qwen2.5, the largest improvement reaches an absolute 50\% points (Development text), and the smallest improvement is 12.6\% points (Creative icons).

We also achieve strong performance on ScreenSpot and ScreenSpot-v2 shown in Table~\ref{tab:screenspot-full}. On both benchmarks, our method attains the highest overall accuracy among all reported approaches, with 85.8\% and 87.8\% respectively, using the fewest training samples. Across the two benchmarks, apart from the Mobile category, we achieve the best performance among the reported methods in the other two major categories as well. This demonstrates the effectiveness of our coarse-to-fine strategy. Although GUI-C$^2$ is primarily designed to address challenging clicking targets such as those in ScreenSpot-Pro, its well-designed policy also improves the ability to handle general, relatively simpler clicking targets. A deeper analysis shows that GUI-C$^2$ performs slightly worse in the Mobile category (though it still reaches 85.5\% and 88.2\%), while showing stronger proficiency in Desktop and Web-style clicking tasks.

\subsection{Ablation Study}

\textbf{Strategy generalization.} We train a 7B model using the same data filtering and method strategy. As shown in Table~\ref{tabpro7b}, our method achieves state-of-the-art performance compared to methods using the same base model, reaching 50.8\%, which demonstrates the generality of our approach.

\textbf{Effectiveness of coarse-to-fine policy.} We first set the number of stages that can invoke the crop tool to one and select a crop ratio of 40\% to ensure that the final crop region approximately matches the result of the two-stage coarse-to-fine refinement. Under this setting, although inference time decreases, accuracy drops by a relative 9.8\%, demonstrating the effectiveness of our policy in improving accuracy. Furthermore, we test a scenario with no tool use at all. Inference time decreases substantially due to the reduction in stages, but accuracy correspondingly drops by a relative 19.4\%, indicating that the tool brings critical improvements shown in Table~\ref{tabab}.

\textbf{Effectiveness of difficulty-aware weighting.} We remove the difficulty-based adjustment from the training process while keeping the same training samples for a fair comparison. The result shows a relative accuracy drop of 4.8\%, demonstrating that increasing the influence of samples that the baseline model finds more difficult indeed enables GRPO to utilize training data more efficiently and learn to solve challenging problems.

\begin{figure}
    \centering
    \vspace{-12pt}
    \includegraphics[width=0.9\linewidth]{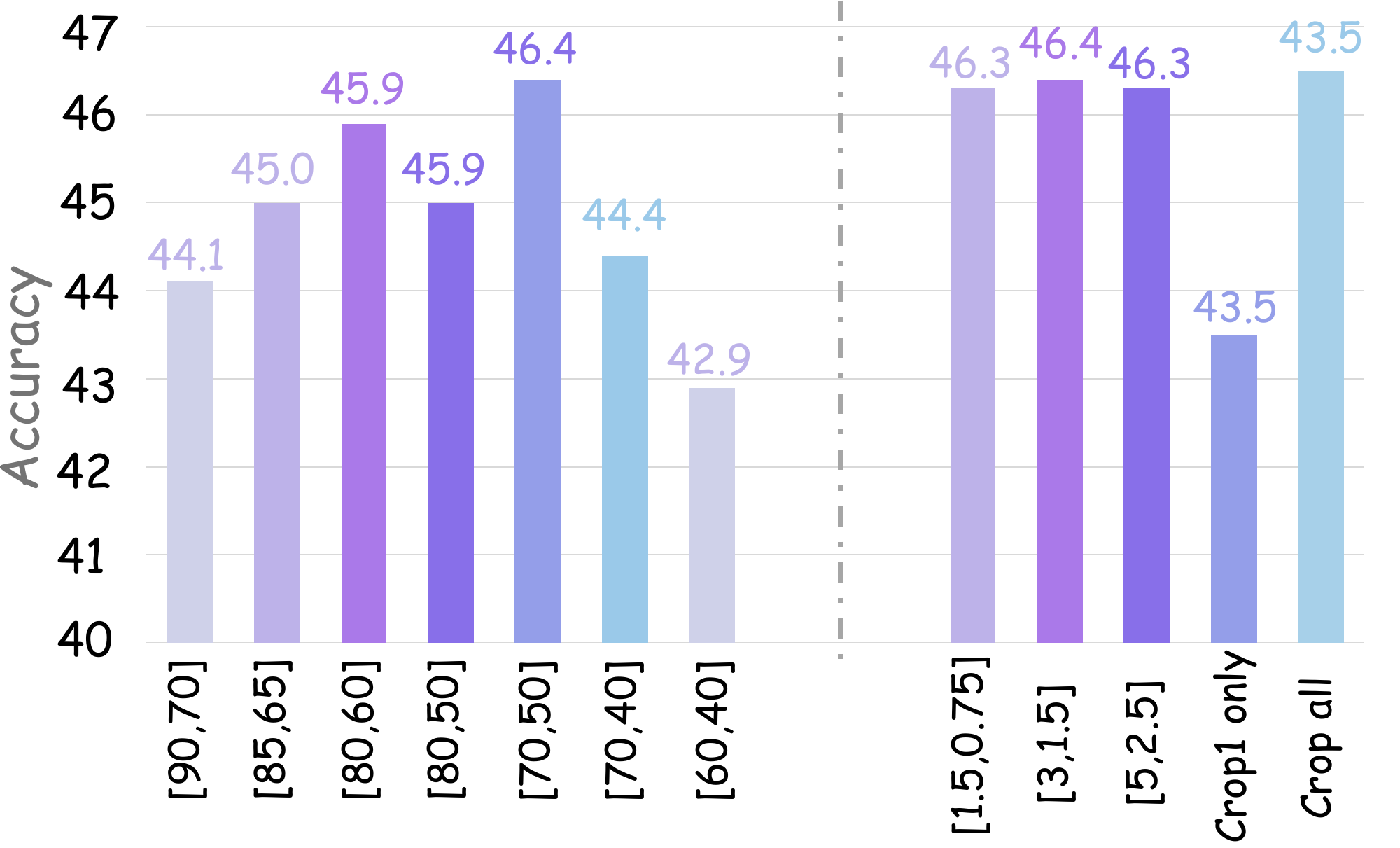}
\vspace{-6pt}
    \caption{Hyperparameter study on ScreenSpot-Pro.}
    \label{figh1}
    \vspace{-16pt}
\end{figure}

\textbf{Effectiveness of simplified model decision-making.} Under comparable conditions without enforcing a thinking process, we remove the area threshold that assists in deciding whether to invoke the crop tool, and instead let the model decide on its own whether to click directly or to crop. The crop ratio is kept consistent with GUI-C$^2$. As a result, inference time increases substantially, while accuracy drops by a relative 5.2\%. This indicates that simplifying the decision-making is a wise choice: it not only reduces the model's inference burden but also lowers the learning difficulty, ultimately leading to improved accuracy. We further allowed the model to decide not only the action but also the crop ratio. As a result, inference time increased further, while accuracy remained almost unchanged.

\textbf{Effectiveness of fixed crop area.} We set the crop ratio to be proportionally adjusted based on the bbox size predicted by the model at each stage, while keeping the crop center unchanged. Under this setting, model performance degrades by a relative 4.6\%, demonstrating that under the coarse-to-fine strategy, a fixed crop area yields better results. The reason may be that the 3B model itself has relatively weak reasoning ability. The coarse-to-fine paradigm already inherently includes the ability to determine the appropriate crop size based on the need. With the removal of the thinking stage to accelerate inference, introducing variable crop ratio adds complexity to the learning process, ultimately leading to negative effects.

\begin{figure}
    \centering
    \vspace{-12pt}
    \includegraphics[width=0.9\linewidth]{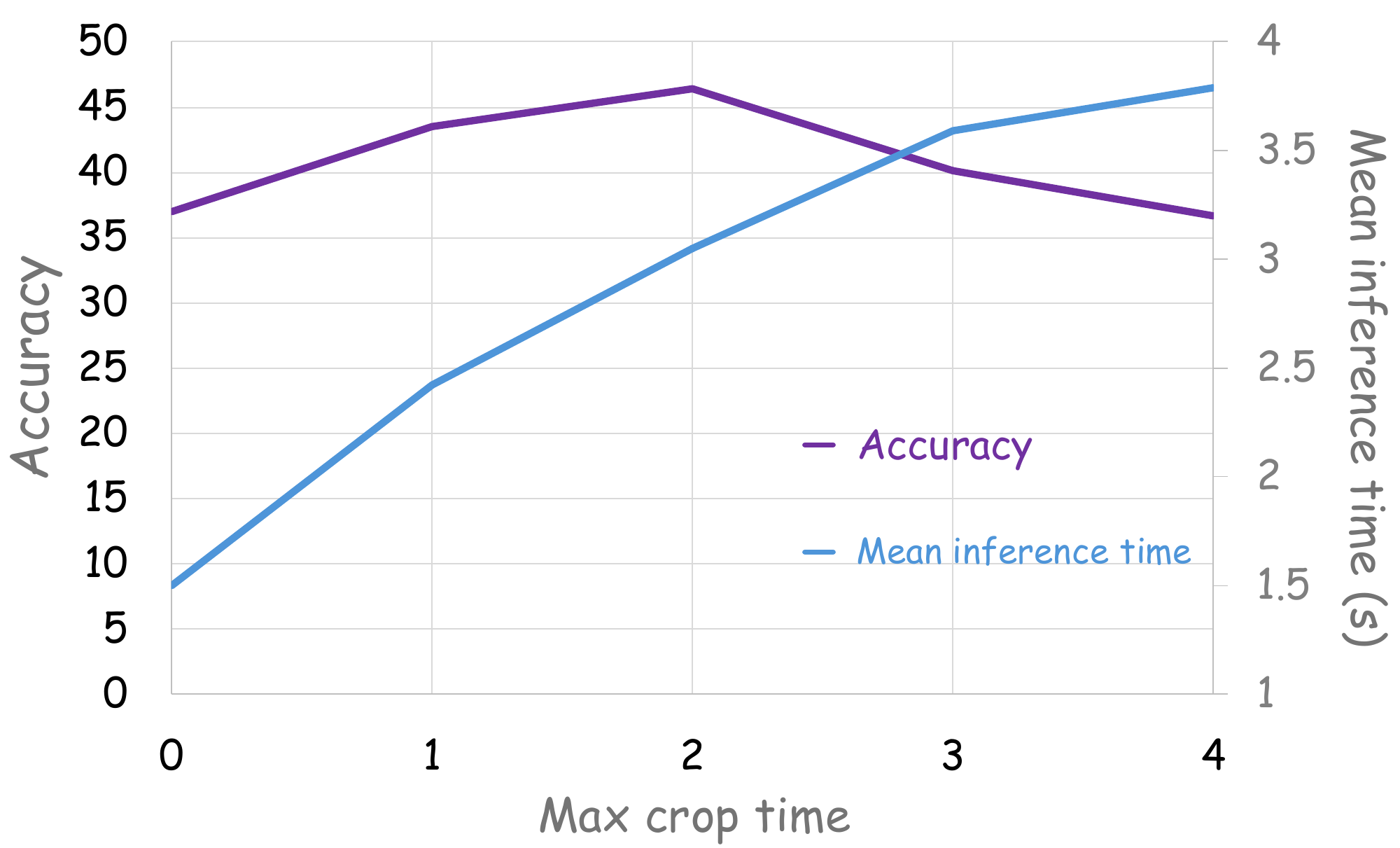}
\vspace{-6pt}
    \caption{Comparison of the impact of different maximum allowable crop stages on GUI-C$^2$-3B.}
    \label{figh2}
    \vspace{-16pt}
\end{figure}

\subsection{Hyperparameter Study}

As shown in Fig.~\ref{figh1} and Fig.~\ref{figh2}, to deeply investigate the impact of different crop ratios $\gamma_1$ and $\gamma_2$ and different area thresholds $\tau_1$ and $\tau_2$ on model performance, we conduct extensive experiments. Under the conditions of $\tau_1=0.03$ and $\tau_2=0.015$, the model achieves its best performance of 46.4\% with $\gamma_1=0.7$ and $\gamma_2=0.5$. As the crop area decreases, we observe a substantial performance drop when $\gamma_2=0.4$, indicating that smaller crop areas are not always better. This may be because overly small cropping loses too much context, thereby impairing the model's spatial localization judgment. Furthermore, to investigate how many crop stages are most suitable for the current model, we evaluate performance under different numbers of crop stages. The results show that as the number of stages increases, inference time steadily rises, while model achieves the optimal performance when the maximum number of allowed crop stages is set to 2.

\section{Conclusion}

In this work, we address two key limitations of agentic RL frameworks for GUI grounding. We propose GUI-D pipeline, which selects training samples with higher value for GRPO, measures each sample’s difficulty for the baseline model through testing, and dynamically adjusts training based on these scores to improve efficiency. Further, we introduce GUI-C² framework, which adopts a coarse-to-fine tool invocation strategy for more effective cropping, while simplifying model decision-making without explicit thinking. Our method achieves 46.4\% accuracy on ScreenSpot-Pro for the 3B model and 50.8\% for the 7B model.

\section*{Limitations}

At the strategy level, although our method simplifies the model's decision-making difficulty using area thresholds and fixed crop ratios, achieving promising experimental results and inference efficiency, it makes the model's output performance dependent on pre-specified hyperparameters. This necessitates subsequent hyperparameter experiments to explore the optimal parameter combinations. Exploring how to enable small-parameter models to autonomously adjust cropping to achieve an optimal balance between context and field of view without incurring excessive additional inference time is our future direction. At the data level, our data filtering pipeline still requires further exploration. Our training is conducted only on samples where the eight clicks are neither all incorrect nor all correct, meaning that we ignore the value of samples with eight complete failures. These samples are more complex than those in the current dataset. Leveraging these more difficult samples to provide effective training may further improve accuracy. Moreover, our data filtering pipeline overemphasizes the 8-click success rate while neglecting class balance, an aspect that needs to be addressed in future work.

\section*{Ethics Statement}

None of the designs involved in our work raise ethical concerns. Also, all data we use are sourced from open-source platforms or widely adopted datasets in the field, and do not involve any privacy exposure issues.

\bibliography{anthology,custom}
\bibliographystyle{acl_natbib}

\appendix

\section{Detailed Definition of Difficulty Score}
\label{apa}

For each remaining sample, GUI-D computes a scalar difficulty score from rollout success, target size, localization error, click dispersion, and parsing failures. 
Let
\begin{equation}
r_i=\frac{c_i}{K}
\end{equation}
be the 8-click success rate. For the ground-truth bounding box $b_i=(x_1,y_1,x_2,y_2)$, we define
\begin{equation}
A_i=(x_2-x_1)(y_2-y_1),
\end{equation}
\begin{equation}
s_i=\min(x_2-x_1,\;y_2-y_1),
\end{equation}
where $A_i$ is the normalized area and $s_i$ is the shorter side length. Since some click targets, such as text-type targets, have a relatively large area but are elongated with width much greater than height, making them still very difficult to click, we adopt an engineering refinement when computing the difficulty score regarding target size. Specifically, we consider not only the area but also the length of the shortest side.

For a predicted point $p=(x,y)$, its distance to the ground-truth box is
\begin{equation}
d(p,b_i)
=
\sqrt{
\Delta_x^2+\Delta_y^2
},
\end{equation}
where
\begin{equation}
\Delta_x=\max(x_1-x,0,x-x_2),
\end{equation}
\begin{equation}
\Delta_y=\max(y_1-y,0,y-y_2).
\end{equation}
Let $\mathcal{P}_i$ be the set of valid parsed click points among the $K$ rollouts. 
The mean localization error is
\begin{equation}
\bar{d}_i
=
\frac{1}{|\mathcal{P}_i|}
\sum_{p\in\mathcal{P}_i}
d(p,b_i),
\end{equation}
and the click dispersion is
\begin{equation}
\rho_i
=
\sqrt{
\frac{1}{|\mathcal{P}_i|}
\sum_{p\in\mathcal{P}_i}
\|p-\bar{p}_i\|_2^2
},
\end{equation}
\begin{equation}
\bar{p}_i=
\frac{1}{|\mathcal{P}_i|}
\sum_{p\in\mathcal{P}_i}p.
\end{equation}
If no valid point is parsed, we set $\bar{d}_i=1$ and $\rho_i=1$. 
Let $m_i$ denote the ratio of unparsed rollouts.

We normalize the difficulty factors as
\begin{align}
g_i^{\mathrm{fail}} &= 1-r_i, \\
g_i^{\mathrm{dist}} &= \min\left(1,\frac{\bar{d}_i}{0.20}\right), \\
g_i^{\mathrm{short}} &= 1-\min\left(1,\frac{s_i}{0.12}\right), \\
g_i^{\mathrm{area}} &= 1-\min\left(1,\frac{\sqrt{A_i}}{0.22}\right), \\
g_i^{\mathrm{disp}} &= \min\left(1,\frac{\rho_i}{0.25}\right), \\
g_i^{\mathrm{miss}} &= m_i.
\end{align}
The raw difficulty score is then computed as
\begin{equation}
\begin{split}
\tilde{q}_i = &\mathrm{clip}_{[0,1]} \big(
0.40 g_i^{\mathrm{fail}} + 0.20 g_i^{\mathrm{dist}} + 0.15 g_i^{\mathrm{short}} \\&
+ 0.10 g_i^{\mathrm{area}} + 0.10 g_i^{\mathrm{disp}} + 0.05 g_i^{\mathrm{miss}} \big).
\end{split}
\end{equation}
Since different platforms may have different UI layouts and visual complexity, we apply platform-wise percentile normalization:
\begin{equation}
q_i
=
\frac{
\mathrm{rank}_{j\in\mathcal{D}_{p_i}}(\tilde{q}_i)
}{
|\mathcal{D}_{p_i}|-1
},
\end{equation}
where $\mathcal{D}_{p_i}$ denotes samples from the same platform as sample $i$. 
The normalized score $q_i\in[0,1]$ is used as the final difficulty coefficient.

\begin{figure*}
    \centering
    \vspace{-8pt}
    \includegraphics[width=0.9\linewidth]{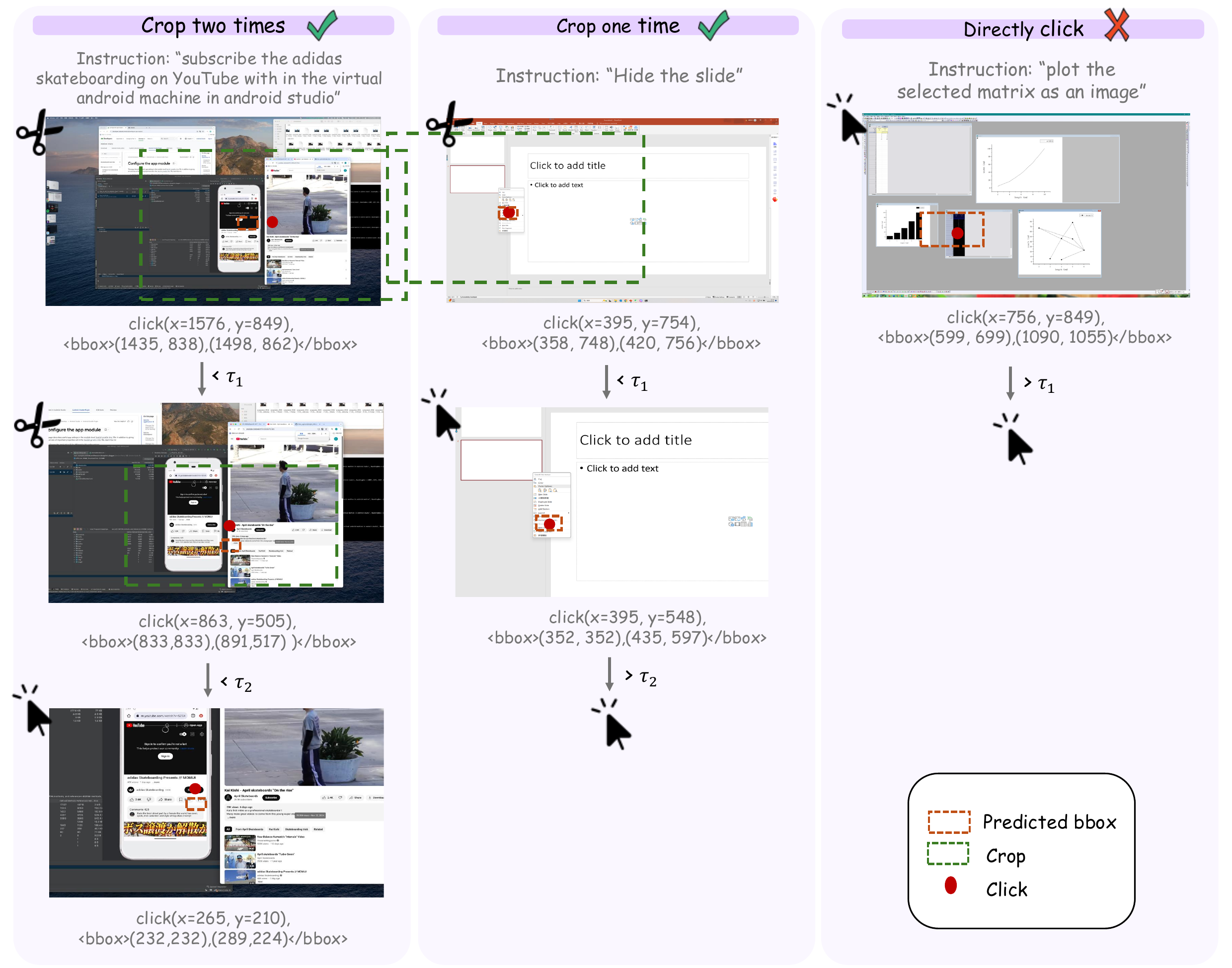}
\vspace{-5pt}
    \caption{Visualization analysis. From left to right, the three examples show results from our testing under two-crop, one-crop, and direct-click actions, respectively. Among them, the results with two crops and one crop are correct, while the result with direct click is incorrect.}
    \label{figv}
    \vspace{-8pt}
\end{figure*}

\section{Supplementary Explanation for GUI-C$^2$}

Our cropping design philosophy centers on the click point output by the model, performing an isometric crop relative to the original image. A crop ratio of 0.7 means that ideally both the width and height become 0.7 times those of the original screenshot. However, when the center point is near the boundary, the cropped region may extend beyond the original screenshot. In such cases, the cropped area is not artificially padded; instead, only the portion that falls within the crop box is retained. Consequently, the cropped image does not always preserve the same aspect ratio as the original. This design avoids errors introduced by artificial padding and reduces overall framework complexity. Furthermore, the second-stage crop ratio is applied to the first-stage cropped image. Thus, a crop ratio of 0.5 does not mean that the remaining area after the second crop maintains a 0.5 ratio relative to the original screenshot; rather, it is necessarily smaller than 0.5 and is most likely 0.35 of the original. This ensures sufficient cropping intensity and prevents excessive redundant information from being retained after cropping.

Since the IoU formula is too fundamental, we do not present it in the main text
\begin{equation}
    \mathrm{IoU}=\frac{\text{Area of Overlap}}{\text{Area of Union}}.
\end{equation}

\section{Supplementary Experimental Details}
\label{apc}

\textbf{Training.} For each training sample, we sample $G=8$ completions in GRPO rollouts. We set $\lambda_d=0.4$, $\alpha_{\min}=0.9$ and $\alpha_{\max}=1.25$. This gives a mild preference to harder samples while avoiding excessive emphasis on noisy cases. And we set $\tau_1=0.03$, $\tau_2=0.015$, $\gamma_1=0.80$, $\gamma_2=0.60$, $\beta^{+}=0.5$, and $\beta^{-}=0.8$. $\beta$ in $R_{click}$ is set to 2.0 in training. The maximum prompt length is 8192 tokens and the maximum completion length is 128 tokens. We set temperature $1.0$ for training-time sampling and set the KL coefficient to $0$ in order to encourage exploration. During training, we adopt an engineering optimization: we enforce two-stage cropping for the first 300 steps as a cold start to prevent the model from becoming too unstable during multi-stage complex training. All the prompts used are shown below, training and evaluation use identical prompts. The training settings for 7B and 3B are kept completely identical.

\paragraph{Inference.} At inference time, GUI-C$^2$ uses deterministic decoding without explicit chain-of-thought generation, set `do\_sample = False' to encourage the reproducible answers. The model outputs only the executable action line. We cap generation at 128 new tokens to avoid long malformed responses. The refinement procedure follows the same deterministic area-gated policy as in training: the model first predicts on the full screenshot, then optionally performs one or two click-centered crops. The final prediction is the deepest valid stage output.

\textbf{Ablation study.} In the ``adaptive crop ratio'' setting, to ensure consistency with GUI-C², we do not allow the model to autonomously decide the crop ratio. Instead, we set it reasonably as an isometric scaling factor. Specifically, the crop ratio is adjusted proportionally based on the size of the target bbox predicted by the model: the smaller the predicted bbox, the smaller the crop area; the larger the predicted bbox, the larger the crop area. This enables more flexible crop region selection under comparable conditions
\begin{equation}
    \mathrm{s}
=
\mathrm{s_min}
+(\mathrm{s_max}-\mathrm{s_min})\sqrt{\frac{\mathrm{a}}{\tau}},
\end{equation}
where $s$ represents the crop ratio. For the predicted bbox at the current stage $b=(x_1,y_1,x_2,y_2)$, the current screenshot size is $W \times H$, and the crop trigger threshold is $\tau$. The normalized area of the bounding box is
\begin{equation}
    \mathrm{a}=\frac{(x_2-x_1)(y_2-y_1)}{WH}.
\end{equation}

This means that when the predicted area is very close to the threshold, the crop size takes the set maximum value $s_{max}$, and when the predicted area is extremely small, the ratio naturally approaches the set minimum value $s_{min}$. The purpose of setting a maximum and minimum value is to keep the cropping within a controllable range without excessively altering the model's overall strategy. In training, we set the first stage $s^{(1)}_{max}=0.85$, $s^{(1)}_{min}=0.7$, the second stage $s^{(2)}_{max}=0.65$, $s^{(2)}_{min}=0.4$, when a format error or parsing error occurs, it falls back to 0.7 and 0.5.

Correspondingly, in the experiments named ``w/ self-action-decide'' and ``w/ self-action-ratio-decide'', we configure the former such that the model must decide whether to crop on its own, rather than relying on threshold comparison based on the predicted bbox. However, this experiment still fixes the crop ratios at $\gamma_1=0.80$ and $\gamma_2=0.60$, and conducts testing under the same setting. In the latter, the model decides not only whether to crop but also the crop ratio. Apart from these settings, we keep all other configurations identical to GUI-C², and neither model is forced to output <think> content in the format.

All ablation experiments are conducted with training, keeping all settings identical except for the ablated module, and they are set to $\gamma_1=0.70$, $\gamma_2=0.50$, $\tau_1=0.03$ and $\tau_2=0.015$ in inference. All inference time measurements were obtained by running the complete ScreenSpot-Pro on 6 NVIDIA 4090 GPUs and computing the average inference time per sample, the time for model loading and data reading was not included.

\textbf{Hyperparameter study.} ``Crop1 only'' means that only the first-stage crop is performed when the condition is met. ``Crop all'' means that all samples undergo both stages of cropping. All experiments were conducted under the same model weights.

Ablation study and hyperparameter study are all conducted on GUI-C$^2$-3B.

\section{Visualization Analysis}

To clearly demonstrate the effectiveness of our strategy, we select representative results for each of the three actions: two crops, one crop, and direct click. As shown in Fig.~\ref{figv}, the ScreenSpot-Pro test data indeed presents high clicking difficulty, with extremely small targets. Both two-crop and one-crop actions help the model narrow down the region through reasonable cropping. For the direct-click example, however, the sample is highly misleading, causing the model to incorrectly predict the target bbox and thus proceed directly to the clicking stage. This incorrect prediction leading to a direct click is a current limitation of our method. Improving prediction accuracy remains our future direction.

\section{Statistics of The Filtered Training Set}

As shown in Table~\ref{tabdata}, our dataset source selection and data distribution refer to previous work.

\begin{promptbox}[Stage 1 — System Prompt]
\small
You are a GUI grounding agent. Given a screenshot and a task, click the target element and predict its bounding box.\\

The screen resolution is \{w\}x\{h\}.\\

Output format: click(x=<px>, y=<py>), <bbox>(<x1>,<y1>),(<x2>,<y2>)</bbox>\\

where (x, y) is the click coordinate in pixels, and (x1, y1), (x2, y2) are the top-left and bottom-right corners of the target element's bounding box in pixels.\\

Output ONLY the action line. Do not add any explanation.

\end{promptbox}

\begin{promptbox}[Stage 1 — User Prompt]
\small
(Constructed by chat template, consists of an image + the task text from the dataset)
\end{promptbox}

\begin{promptbox}
[Stage 2/3 (Cropped Region) — System Prompt]
\small
You are a GUI grounding agent. The image shown is a zoomed-in region of a screenshot.\\

The region resolution is \{w\}x\{h\}.
Click the target element precisely.\\

Output format: click(x=<px>, y=<py>), <bbox>(<x1>,<y1>),(<x2>,<y2>)</bbox>\\

Output ONLY the action line.
\end{promptbox}

\begin{promptbox}[Stage 2/3 (Cropped Region) — User Prompt]
\small
Task: \{task\}\\

This image is a zoomed-in crop from the original screenshot. Use the task above to identify the correct target element and click it precisely in this cropped view.
\end{promptbox}

\section{Supplementing Experiments with Different Reward Function Parameters}

To more deeply explore the impact of reward function hyperparameter settings on experimental results, we thoroughly investigate the accuracy of GUI-C²-3B trained under different parameter configurations on ScreenSpot-Pro as shown in Table~\ref{tabrf}. During training, all settings other than the parameters themselves are kept completely consistent.
\begin{table}
    \centering
    \scalebox{0.68}{
    \begin{tabular}{cccccc}
    \hline
         & Mobile & Web & Windows & Linux & MacOS\\
         \midrule
        Source & UI-BERT & SeeClick & OS-Atlas & OS-Atlas & OS-Atlas\\
        Size for 3B & 359 & 2262 & 1577 & 246 & 180\\
         Size for 7B & 276 & 2593 & 1529 & 246 & 180\\
        \bottomrule
    \end{tabular}
}
    \caption{Statistics of training set.}
    \label{tabdata}
\end{table}

\begin{table}
    \centering
    \begin{tabular}{cccc}
    \hline
      $\lambda_{\mathrm{click}}$  & $\lambda_{\mathrm{iou}}$  & $\lambda_{\mathrm{fmt}}$  & Avg.\\
      \midrule
      0.6 & 0.3  & 0.1& 46.4\\
        0.5 & 0.4 & 0.1 & 45.7\\
        0.7 & 0.2 & 0.1& 46.1\\
   \bottomrule
    \end{tabular}
    \caption{Performance comparison of different reward function parameters on ScreenSpot-Pro.}
    \label{tabrf}
\end{table}

\end{document}